\documentclass[10pt, a4paper, oneside]{article}

\usepackage{array}
\usepackage{graphicx}
\usepackage{multirow}
\usepackage{amsmath}

\usepackage{amssymb}
\usepackage{siunitx} 
\usepackage{enumitem}
\usepackage{hyperref} 
\usepackage{url} 
\usepackage[T1]{fontenc}
\usepackage{draftwatermark} 

\usepackage{nameref}
\usepackage{booktabs}

\usepackage{subfigure}
\usepackage{multimedia}

\title{Automatic Segmentation of Aircraft Dents in Point Clouds}

\author{Pasquale Lafiosca$^a$ \and
        Ip-Shing Fan$^a$ \and Nicolas P. Avdelidis$^a$
}

\date{\small{$^a$Integrated Vehicle Health Management Centre, Cranfield University, United Kingdom}
}

\begin{document}

\maketitle

\SetWatermarkText{Article in Press} 
\SetWatermarkScale{3} 

\section{Abstract}
Dents on the aircraft skin are frequent and may easily go undetected during airworthiness checks, as their inspection process is tedious and extremely subject to human factors and environmental conditions.
Nowadays, 3D scanning technologies are being proposed for more reliable, human-independent measurements, yet the process of inspection and reporting remains laborious and time consuming because data acquisition and validation are still carried out by the engineer.
For full automation of dent inspection, the acquired point cloud data must be analysed via a reliable segmentation 
algorithm, releasing humans from the search and evaluation of damage.
This paper reports on two developments towards automated dent inspection. The first is a method to generate a synthetic dataset of dented surfaces to train a fully convolutional neural network. The training of machine learning algorithms needs a substantial volume of dent data, which is not readily available. Dents are thus simulated in random positions and shapes, within criteria and definitions of a Boeing 737 structural repair manual.
The noise distribution from the scanning apparatus is then added to reflect the complete process of 3D point acquisition on the training.
The second proposition is a surface fitting strategy to convert 3D point clouds to 2.5D. This allows higher resolution point clouds to be processed with a small amount of memory compared with state-of-the-art methods involving 3D sampling approaches.
Simulations with available ground truth data show that the proposed technique reaches an intersection-over-union of over $80\%$. Experiments over dent samples prove an effective detection of dents with a speed of over $\SI{500000}{}$ points per second.

\section{List of Abbreviations}
\begin{table}[h]
\begin{tabular}{ll}
BCE & Binary cross entropy \\
BN & Batch normalisation \\
CNN & Convolutional neural network \\
FCN & Fully convolutional network \\
IoU & Intersection over union \\
MLP & Multilayer perceptron \\
R-CNN & Region based convolutional neural network  \\
ReLU & Rectifier linear unit \\
SRM & Structural repair manual\\
\end{tabular}
\label{tab:abbr}
\end{table}

\section{Introduction}
Aircraft visual inspections are costly, time-consuming and sometimes hazardous procedures, yet essential for the airworthiness assessment. Research also reveals the subjective nature of engineers' evaluations~\cite{CAA_CAP716, CAA_CookThesis}.
Machine vision, on the other hand, is progressing at an incredible rate, surpassing humans in many occasions~\cite{coffey2018machine}, with great potential to increase both efficiency and efficacy of inspections.

Nevertheless, the automation of aircraft visual inspections is still very limited.
Automation of dent inspections, in particular, is challenging due to the nature of the damage itself. A dent is smooth, without clear boundaries and difficult to identify even by trained engineers. In addition, its evaluation demands to collect measures as prescribed by the structural repair manual (SRM). Hence, 2D pattern recognition of monocular images is useless for the scope, because metric information is absent.
Even if it solves a part of the problem producing metric data, the use of 3D scanning technologies leaves the operator with a substantial amount of raw data, or \textit{point clouds}, from which information has to be extracted manually.
As such, the correct labeling of each point as belonging to a dent or not, also known as \textit{semantic segmentation}, becomes essential for comprehensive automation.

While the use of \textit{ad hoc} rule-based algorithms is anachronistic, the actual application of machine learning technologies is not straightforward, mainly for two reasons. First, the lack of a publicly accessible, \textit{labelled} and substantial dent dataset required for the training and validation. Secondly, because dent segmentation is very different from the segmentation of general objects, as local features are not particularly distinctive over the aircraft skin.

This study addresses the problem of semantic segmentation in aircraft dent inspections, proposing two concepts.
First, an innovative process to build a \textit{virtual} dataset: dented surfaces are randomly generated according to the model explained below and mixed with \textit{noise signals} extracted from the chosen 3D scanning apparatus, adding robustness during training.
The dataset generated in such way is accompanied by its \textit{ground truth data}, i.e. the true classification for each one of the points, labelled as dented or not dented. Being virtually generated, this classification is exact and not subject to measuring or evaluation errors, and provides the basis for error backpropagation during the training of a machine learning algorithm.

Secondly, the problem dimensionality is reduced by the employment of an effective surface fitting strategy, where 3D points are reduced to 2D by projection onto the fitted bivariate quadric surface. Then, for each point is possible to calculate a residual or distance from the fitted surface.
The 2D residuals are then fed to a fully convolutional network (FCN) inspired from U-Net~\cite{unet2015} for quick, lightweight and accurate surface segmentation.
Simulations with available ground truth data show that the proposed technique reaches an intersection-over-union (IoU) (or \textit{Jaccard index}) of over $80\%$. Experiments over dent samples prove the effective segmentation of dents with a speed of over $\SI{500000}{}$ points per second.

\section{Previous work}
The general problem of segmentation has been extensively discussed in the literature, while only a few articles dealing with dent segmentation. Methods can be classified as 2D or 3D, according to their input dimensionality, and considering whether they employ machine learning or \textit{ad hoc} algorithms based on handcrafted features and rules~\cite{xie2020linking}. 

Among the latter, region growing has been applied to both 2D and 3D data and uses criteria to combine features of local points, merging them together if they have similar properties.
Jovančević et al.~\cite{jovanvcevic20173d} proposed a method to detect and characterise aircraft dents based on the estimation of normal and curvature values. The defect is characterised after 2D projection and further processing.
As point-wise calculation is generally slow, the processing time reported was $\SI{20}{\second}$ for $\SI{30000}{}$ points, and up to $\SI{120}{\second}$ for denser point clouds (on a PC with 2.4 GHz Core i7 CPU, 8GB RAM).
Other non-AI approaches are, for example, clustering and model fitting~\cite{xie2020linking}.
For all non-AI methods, preprocessing parameters and thresholds affect the algorithm sensitivity and must be tuned accordingly for each scenario. Since the user interaction is continuously required, no proper automation can be achieved.
As such, recent research is mostly oriented towards machine learning.

In~\cite{bouarfa2020towards} a Mask R-CNN~\cite{he2017mask} was used to detect and segment pictures of aircraft dents with average IoU of $36.54\%$. The use of R-CNN was also tested for the sole detection of car dents~\cite{park2020detecting}. In this case light bands were projected on the surface to highlight its deformations.
Although methods processing 2D images via CNNs are the \textit{de facto} standard for image classification, object detection and segmentation~\cite{unet2015}, they come with two main problems when applied to aircraft inspections.
First, \textit{measures} cannot generally be provided as absent in the original input, usually consisting in monocular images. Since measures are the discriminating factor for evaluating defects as allowable damage or not, as prescribed by the SRM, the usefulness of doing an inspection without acquiring them is questionable.
Secondly, while scratches and lightning may be detected reasonably well, the lack of texture and distinctive features over the single-coloured aircraft skin makes the detection of a dent particularly challenging.
Therefore, methods acquiring and processing 3D metric data are more interesting for the scope.

Convolution cannot be applied directly to unordered point clouds~\cite{xie2020linking}. The most straightforward way to apply 2D CNN to 3D data are projection-based approaches. While these allow seamlessly to transfer the same technology to 3D data, projecting in 2D generally causes loss of information and the results are highly dependent on the chosen points of view.
Alternatively, \textit{voxels} address the need of having an ordered structure to apply convolution, creating low resolution, equally-spaced data points. These strategies organise all the volume, including empty space, into some structure~\cite{riegler2017octnet,rethage2018fully}, resulting in high computational cost and memory requirements, often beyond the availability of the average user.
A third and innovative approach is to directly process the point coordinates. PointNet~\cite{qi2017pointnet} pioneered the use of a symmetric function achieving permutation invariance. Also, the input size is often predefined because the data is processed via multilayer perceptron (MLP) layers. As PointNet is not effective at capturing local features, its derivations introduce expensive preprocessing and neighbouring operations~\cite{qi2017pointnet++,liu2019point}.
A substantial number of architectures based on a combination of the above principles is present in the literature, summarised in~\cite{xie2020linking,guo2020deep}, however, a general approach that outperforms the others is yet to be found.

\section{DentNet Segmentation Strategy}
The B737-400 SRM defines a dent as \textit{``a damaged area that is pushed in from its normal contour''} and has \textit{``smooth''} edges~\cite{SRM737400}. However, no formal shape definition is given.

For the scope of this work, a pseudo-realistic synthetic dataset of surfaces is created: dents are simulated in random positions and shapes, following an analytical model within the above SRM criteria and definitions. To increase robustness, the noise distribution from the scanning apparatus is then acquired and added over the synthetic data during the training of the FCN. The network can thus be used for segmentation of real 3D scan data.
The proposed method exploits the nature of surface inspection to deal with the sparsity of point cloud data, making use of a 2.5D strategy. This allows higher resolution point clouds to be processed with a small amount of memory compared with state-of-the-art methods.

\subsection{Virtual Dataset}
As anticipated above, the virtual dataset generation is necessary to overcome the lack of real data.
Curved surfaces are considered with shapes compatible with most of the aircraft areas. These are then virtually dented.

The world $xy$-dimensions of the surface are fixed as $X \times Y$ (metric units), sampled over a grid of $w \times h$ pixels.
For each sample, $xy$-values are fed to a parabola function:
\begin{equation}
 z_s = \alpha x^2 + \beta y^2
\end{equation}
with $\alpha$ and $\beta$ randomly chosen within a certain range, compatible with the curvature of the aircraft surfaces to be inspected. Later, noise from a uniform distribution is added to $xy$-values.

A dent is modeled by:
\begin{equation}\label{eq:dentFunction}
 z_d = \begin{cases}
      -e^{-\frac{1}{1-r^2}}  & \quad \text{if $|r|<1$}\\
      0   \quad & \quad \text{elsewhere}
     \end{cases}
\end{equation}
where $r = \sqrt{x^2 + y^2}$. The function is randomly rescaled in $x$ and $y$ to cover different dent dimensions and proportions, rotated and shifted with respect to the surface.
A dent is added to the surface with a certain probability $p$. Further dents are added to the same surface with exponential decreasing probability.

The depth of the sample is finally obtained as:
\begin{equation}\label{eq:zValues}
z = z_s + \max \left(z_d^{(1)}, \ldots, z_d^{(n)} \right) + z_w 
\end{equation}
where $n$ is the total number of dents and $z_w$ represents a Gaussian white noise distribution, added for extra robustness during training.
$xy$-values and $z$-values are joined together to form the final dented surface, which is then randomly rotated along all the three world axes, covering the possible angles of a real scanning scenario.
The surface here generated is thus represented by 3D coordinates, still ordered into a $w \times h$ matrix that preserves relative proximity of points.

\subsection{Noise Signals}
The noise distribution when scanning real samples is not necessarily Gaussian and depends on the selected scanning apparatus and its 3D reconstruction algorithm.
Although mixing virtual and real data for training is not a new concept in literature~\cite{waferVirtualdataset, tian2018virtualDataset}, the innovative procedure here proposed consists in mixing virtual data with only \textit{noise signals} from real world.

\begin{figure}[h!]
 \centering
 \includegraphics[width=\linewidth]{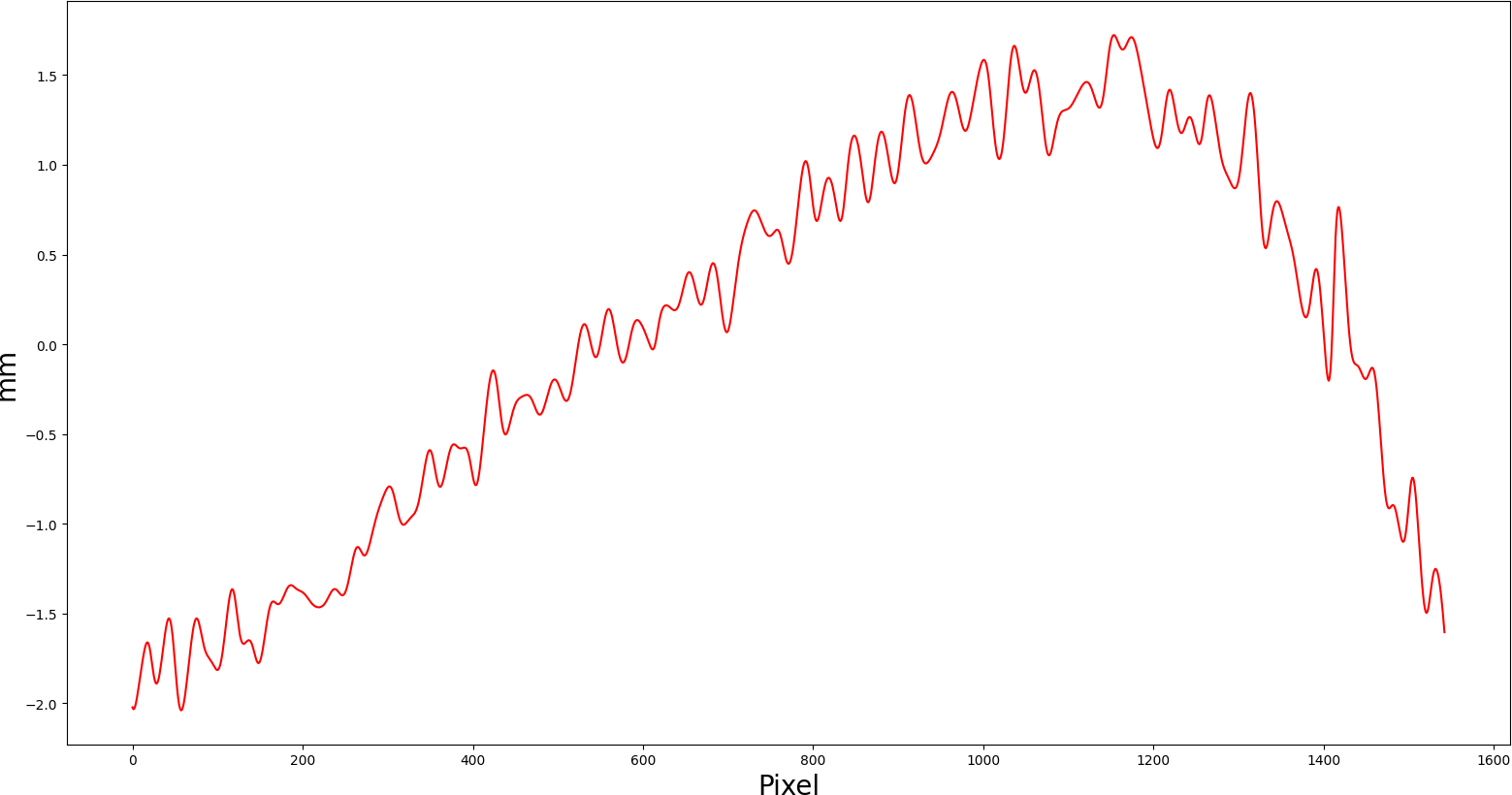}
 \caption{Example of noise distribution of the scanning apparatus.}
  \label{fig:noiseOverPlane}
\end{figure}

The signals are obtained scanning flat a white board several times and in different positions with the 3D scanner. In this case a system based on Fourier transform profilometry~\cite{feng2021calibration, myFTPmethod} was chosen.
An example of its noise distribution over a flat board about $\SI{40}{\centi\meter}$ wide is shown in Fig.~\ref{fig:noiseOverPlane}.

\begin{figure}[h!]
 \centering
 \includegraphics[width=0.9\linewidth]{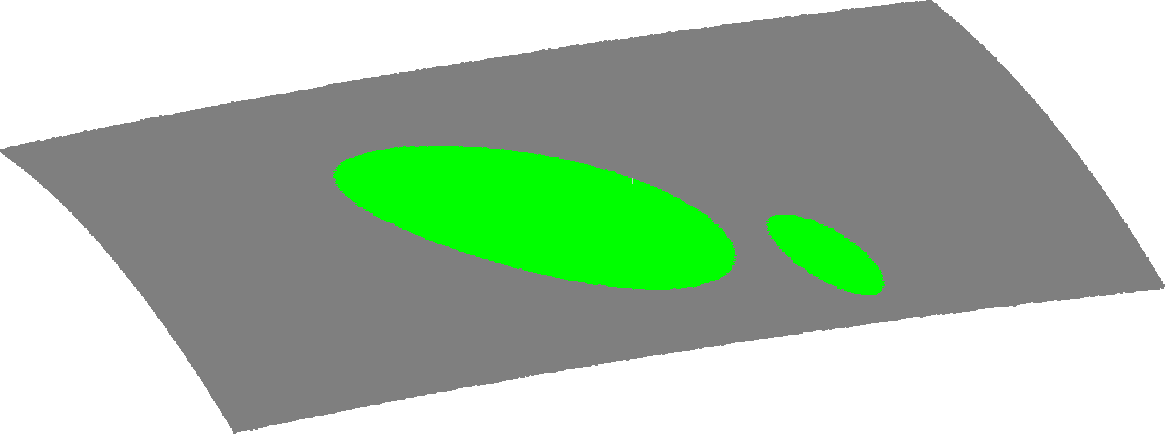}
 \caption{Virtual dented surface with ground truth (dents in green).}
  \label{fig:virtualSurface}
\end{figure}

The acquired signals are augmented via random cropping and flipping, and then added to the $z$ value of Eq.~(\ref{eq:zValues}) before proceeding with preprocessing and training. An example of resulting dented surface is shown in Fig.~\ref{fig:virtualSurface}.

The above strategy allows to produce more plausible data (accompanied by its ground truth), which is used to train a neural network that can eventually translate its knowledge to real dent samples.
Compared to the addition of only Gaussian noise, this strategy leads to a considerable improvement of the predictions when scanning real samples with the same apparatus, as confirmed by the following experiments.

\subsection{Input Preprocessing}
Input 3D data consists in a point cloud, i.e. a list of \textit{xyz} coordinates derived from any 3D scanning device.
For quick and effective segmentation, a bivariate quadratic function is first fitted to the 3D points, then the residuals are calculated as distances along $z$-axis, \textit{de facto} providing 2D data that can be used as input of a FCN.
The data preprocessing pipeline consists in:
\begin{enumerate}[noitemsep,topsep=-0.5em]
 \item Canonical rotation of the surface;
 \item Bivariate quadratic function fitting;
 \item Extraction of the 2D matrix of residuals.
\end{enumerate}
First, the surface is oriented so that its average normal direction $\hat{n}$ is parallel the world direction $\hat{z}$ (coincident with the scanner optical axis), where $\hat{n}$ is calculated by fitting a plane and the rotation to apply is found as:
\begin{equation}
 \mathbf{R} = \mathbb{I} + \left[\hat{z} \times \hat{n}\right]_\times + \left[\hat{z} \times \hat{n}\right]^{2}_\times \frac{1}{1+\hat{z}\cdot\hat{n}}
\end{equation}
where $\mathbb{I}$ is the $3\times3$ identity matrix and $[\ \ ]_\times$ indicates the cross-product skew-symmetric matrix.
In our setup is always $\hat{z}\cdot\hat{n} \neq -1$.

Then, a bivariate quadratic function expressed as:
\begin{equation}\label{eq:quadratic}
 z(x,y) = a + b x + c y + d x^2 + e xy + d y^2
\end{equation}
is fitted to the surface. $z(x,y)$ is general enough to represent the shape of most of the aircraft skin, \textit{neglecting} the potential dents there present and following its baseline.
The residuals (or distances from the fitted surface) of this fitting are thus used as input for the FCN in the form of a $w \times h$ matrix. In such way dents will appear as consistent high-residual areas, which the network should learn to segment comparing its predictions with the ground truth.

Dents do not have distinctive characteristics and the conversion of 3D points to distances from the quadratic function is essential to highlight deformations that the FCN can discriminate. Furthermore, point data is reduced to 2D for quick and lightweight processing.
No expensive neighbouring searching mechanism is used and the preprocessing is reasonably fast, taking on average $\SI{0.4}{\second}$ for $\SI{614400}{}$ points (Intel i7-7500U CPU, 12 GB of RAM).

\subsection{Network Architecture}
U-Net is a FCN that was developed for medical image segmentation~\cite{unet2015}. It is composed by a downsampling part that 
converts the input image to a compressed set of features, similarly to what is done for image classification where these are then linked to labels. However, with U-Net these features are not linked to labels but fed to an upsampling part, symmetric to the previous one, that expands the dimensions back to the original ones, delivering a label for each input value (segmentation).

The proposed network is fully convolutional, inspired from U-Net and with a similar number of trainable parameters, although with some key differences. The encoder part consists in the repeated application of two convolution operations with kernel size $3 \times 3$ and padding size $1$. The first, identified with \textit{Conv}, has stride $1$. The second one, identified with \textit{DownConv}, has stride $2$: it halves the dimensions while doubling the number of channels.
The decoder part has repeated application of \textit{Conv} followed by a transposed convolution layer, or \textit{UpConv}, having kernel size $3 \times 3$ and both padding and output padding set to $1$, doubling the dimensions and halving the number of channels.

\begin{figure}[ht!]
 \centering
 \includegraphics[width=1\linewidth]{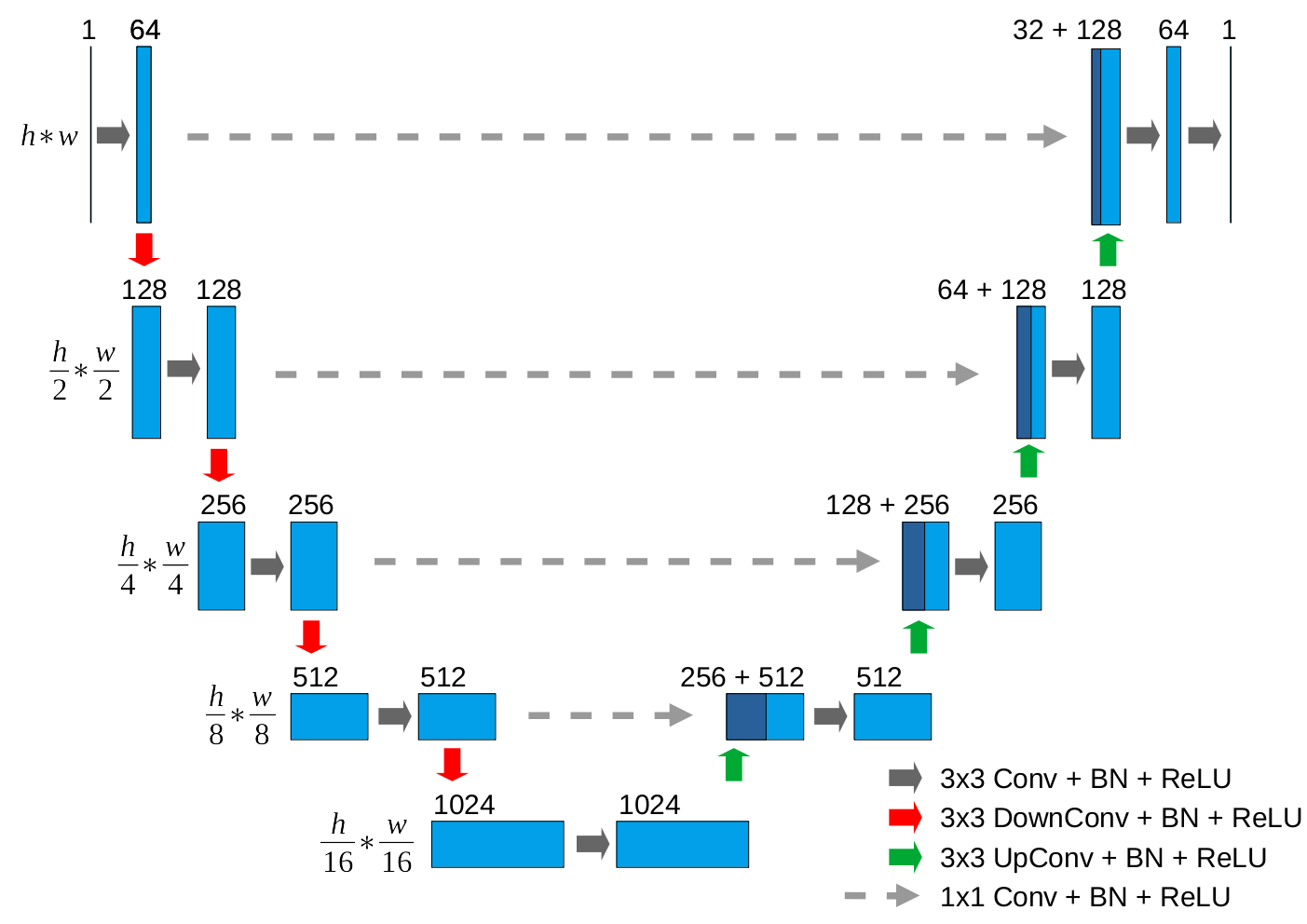}
 \caption{Network diagram.}
  \label{fig:networkArchitecture}
\end{figure}

Each convolution is followed by a batch normalisation (BN), that enables faster and stable training~\cite{santurkar2018does}, and a rectifier linear unit (ReLU) as activation function. In the output layer, the latter is replaced by a sigmoid.
The addition of a dropout layer after each \textit{Conv} was found to impair performance, thus it was omitted.
For better localisation, a skip connection is introduced after each \textit{Conv} in the encoder to increase localisation capabilities. The number of channels is reduced via a $1 \times 1$ convolution and the features are concatenated at the same level in the decoder part.
The network diagram is shown in Fig.~\ref{fig:networkArchitecture}.

The training is performed by using binary cross entropy (BCE) as loss function.
However, the input data may be \textit{unbalanced}: dented areas (positive) are generally much less than non-dented areas (negative). Thus, a balancing weight is estimated from the input data and applied during loss calculation, increasing recall.
The network can accept any input size with dimensions multiple of $16$, nevertheless, the point \textit{density}
(points per $\SI{}{\milli\meter}$) of the test set is expected to be roughly the same as the one in the training set.

\section{Simulations and experiments}\label{sec:experiments}
A dataset of $\SI{30000}{}$ dented surfaces was generated as shown above, each one containing $960 \times 640$ points over a spatial dimension of $\SI[product-units=single]{50 x 33}{\centi\meter}$.
The dataset was split assigning $80\%$ for training and $20\%$ for validation. The network was trained using two Nvidia V100 64GB GPUs and each epoch trained in about $\SI{2}{\hour}$.
The goal was to achieve the maximum IoU over the validation set.

For comparison, the network was trained first with Gaussian noise only ($\sigma = \SI{2}{\milli\meter}$), without the use of noise signals. The confusion matrix is shown in Table~\ref{tab:confusionGaussian}. In this case the network reached a IoU of $92.66\%$. Precision and recall were $94.88\%$ and $97.54\%$, respectively.
This result, however, is affected by the more predictable noise distribution and does not necessarily translate to better performance when the network is used on \textit{real} dent samples, as shown in the experiments below.

\begin{table}[h!]
\centering
\begin{tabular}{llll}
 &                       & \multicolumn{2}{c}{\textit{Prediction}}                       \\ \cline{3-4} 
 & \multicolumn{1}{l|}{} & \multicolumn{1}{l|}{\textbf{Positive}} & \multicolumn{1}{l|}{\textbf{Negative}} \\ \cline{2-4} 
\multicolumn{1}{c|}{\multirow{2}{*}[-0.5em]{\rotatebox{90}{\textit{Actual}}}} &
  \multicolumn{1}{l|}{\begin{tabular}[c]{@{}l@{}}\textbf{Positive}\\ \footnotesize(dented)\end{tabular}} & 
  \multicolumn{1}{c|}{$11.49\%$} &
  \multicolumn{1}{c|}{$0.29\%$} \\ \cline{2-4} 
\multicolumn{1}{c|}{} &
  \multicolumn{1}{l|}{\begin{tabular}[c]{@{}l@{}}\textbf{Negative}\\ \footnotesize(non dented)\end{tabular}} &
  \multicolumn{1}{c|}{$0.62\%$} &
  \multicolumn{1}{c|}{$87.60\%$} \\ \cline{2-4} 
\end{tabular}
\caption{Confusion matrix resulting from training with Gaussian noise only. IoU was $92.66\%$.}
\label{tab:confusionGaussian}
\end{table}

Afterwards, $20$ flat surfaces with a resolution of circa $1800 \times 900$ pixels were acquired with the scanning apparatus and their error values calculated and augmented as for classic 2D images, then the network was trained again with both these noise signals and Gaussian noise ($\sigma = \SI{1}{\milli\meter}$) added to the virtual dents.
The $xy$ noise and all the other parameters were left unchanged.

In this second case, the IoU reached was $84.76\%$, lower, as expected, due to the presence of the irregular noise signals that the network has to model.
Precision and recall were $89.35\%$ and $94.25\%$, respectively, while the accuracy $98.05\%$. The confusion matrix of this second case is shown in Table~\ref{tab:confusionNoise}.
\begin{table}[h!]
\centering
\begin{tabular}{llll}
 &                       & \multicolumn{2}{c}{\textit{Prediction}}                       \\ \cline{3-4} 
 & \multicolumn{1}{l|}{} & \multicolumn{1}{l|}{\textbf{Positive}} & \multicolumn{1}{l|}{\textbf{Negative}} \\ \cline{2-4} 
\multicolumn{1}{c|}{\multirow{2}{*}[-0.5em]{\rotatebox{90}{\textit{Actual}}}} &
  \multicolumn{1}{l|}{\begin{tabular}[c]{@{}l@{}}\textbf{Positive}\\ \footnotesize(dented)\end{tabular}} & 
  \multicolumn{1}{c|}{$10.82\%$} &
  \multicolumn{1}{c|}{$0.66\%$} \\ \cline{2-4} 
\multicolumn{1}{c|}{} &
  \multicolumn{1}{l|}{\begin{tabular}[c]{@{}l@{}}\textbf{Negative}\\ \footnotesize(non dented)\end{tabular}} &
  \multicolumn{1}{c|}{$1.29\%$} &
  \multicolumn{1}{c|}{$87.24\%$} \\ \cline{2-4} 
\end{tabular}
\caption{Confusion matrix resulting from training with both noise signals and Gaussian noise. IoU was $84.76\%$.}
\label{tab:confusionNoise}
\end{table}

\begin{figure}[ht!]
 \centering
 \includegraphics[width=\linewidth]{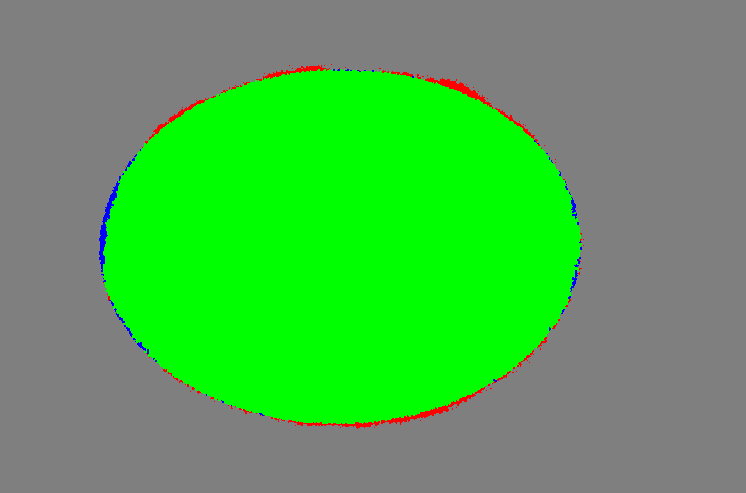}
 \caption{Prediction results over one of the validation samples. True positive are in green, false positive in red and false negative in blue.}
  \label{fig:predictionSimulated}
\end{figure}

The prediction over one of the virtual surfaces in the validation set is shown in Fig.~\ref{fig:predictionSimulated}. 
It presents a rather shallow and wide dent, generally the kind more difficult to detect by humans~\cite{CAA_CookThesis}.
In this case, results can be compared with ground-truth data and, like for human engineers, the boundaries result as the most challenging to identify.
 
\begin{figure}[!ht]
 \centering
  \begin{subfigure}
  \centering
    \includegraphics[width=0.45\linewidth]{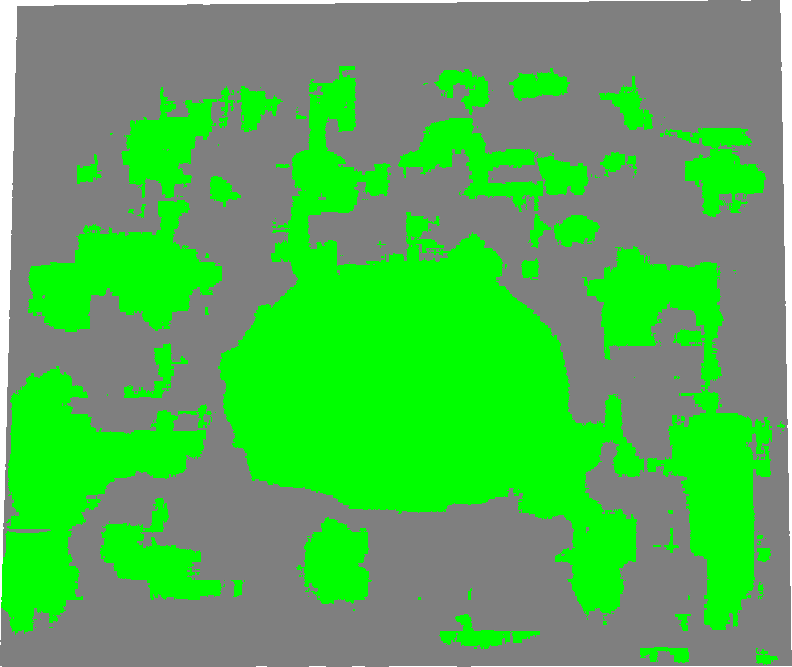}
  \end{subfigure}
  \hspace{0.05\linewidth}
  \begin{subfigure}
   \centering
    \includegraphics[width=0.45\linewidth]{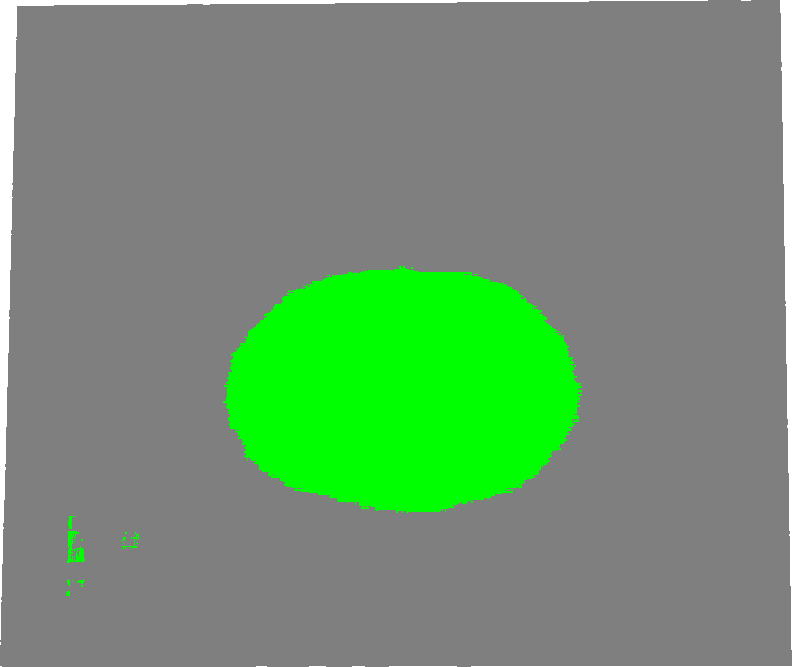}
  \end{subfigure}
  
  \begin{subfigure}
  \centering
    \includegraphics[width=0.45\linewidth]{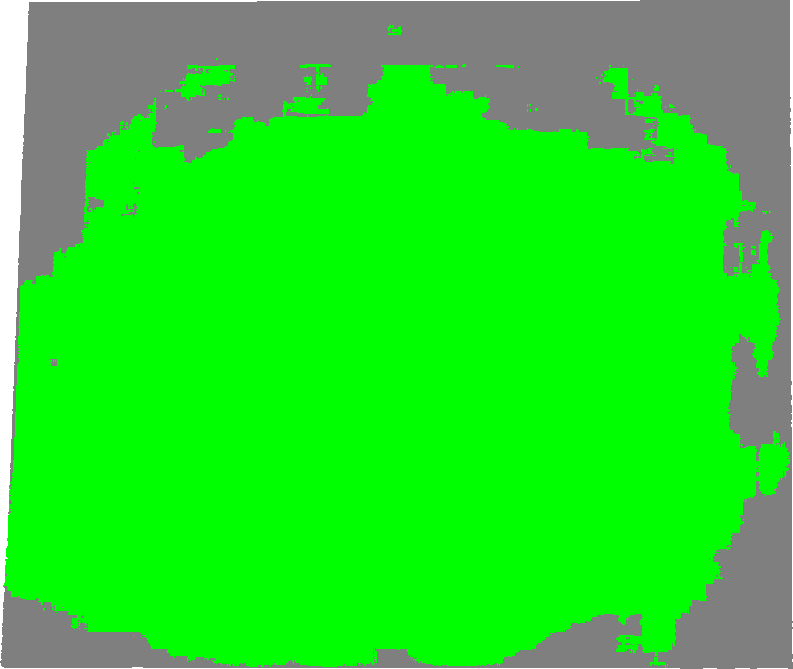}
  \end{subfigure}
  \hspace{0.05\linewidth}
  \begin{subfigure}
   \centering
    \includegraphics[width=0.45\linewidth]{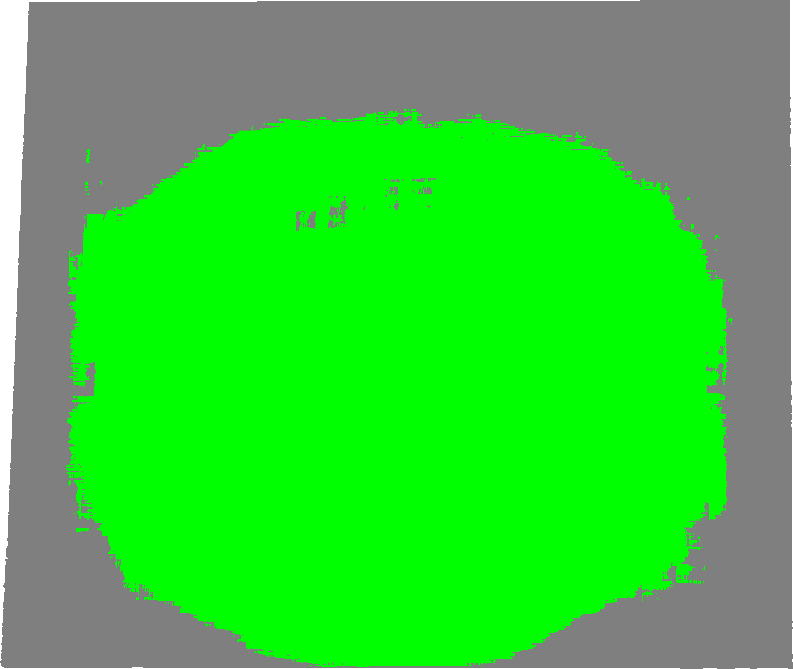}
  \end{subfigure}
  
  \begin{subfigure}
  \centering
    \includegraphics[width=0.45\linewidth]{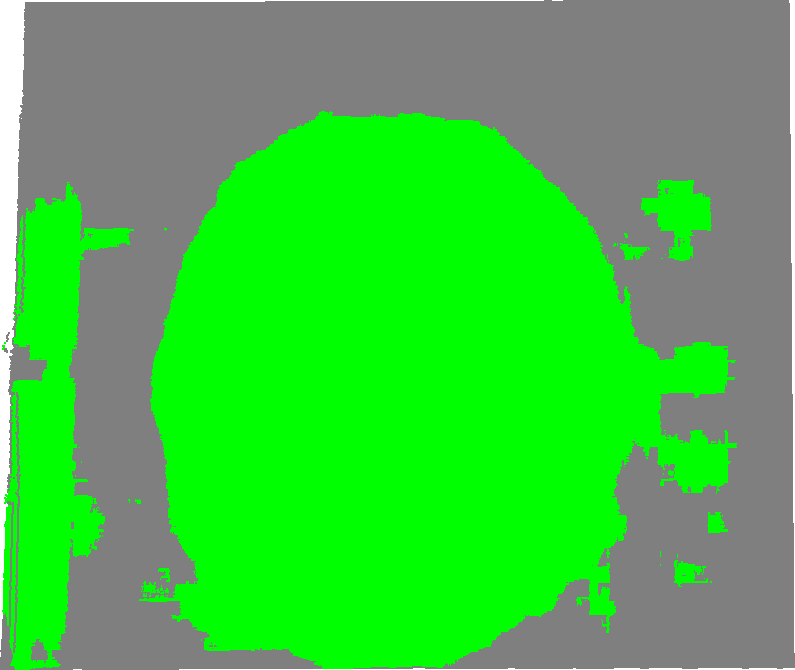}
  \end{subfigure}
  \hspace{0.05\linewidth}
  \begin{subfigure}
   \centering
    \includegraphics[width=0.45\linewidth]{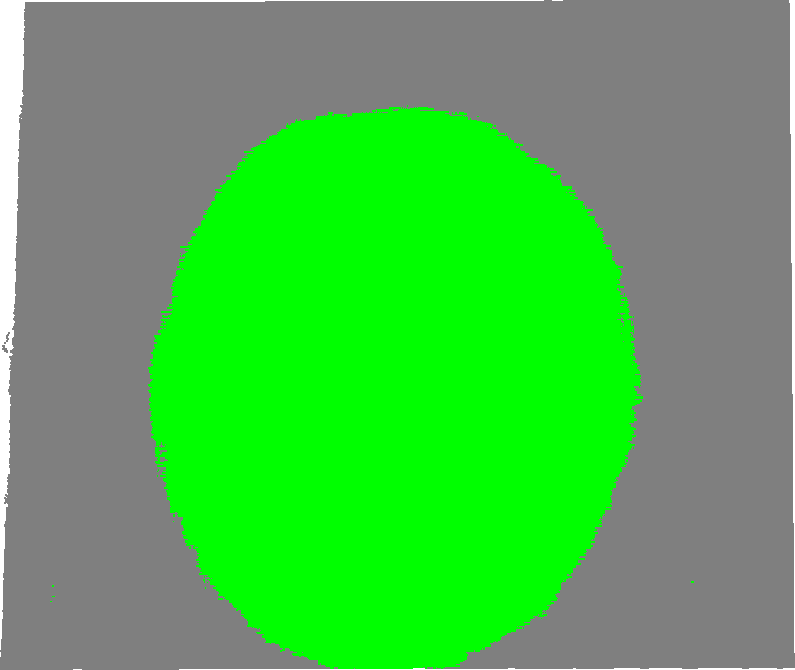}
  \end{subfigure}
  
  \begin{subfigure}
  \centering
    \includegraphics[width=0.45\linewidth]{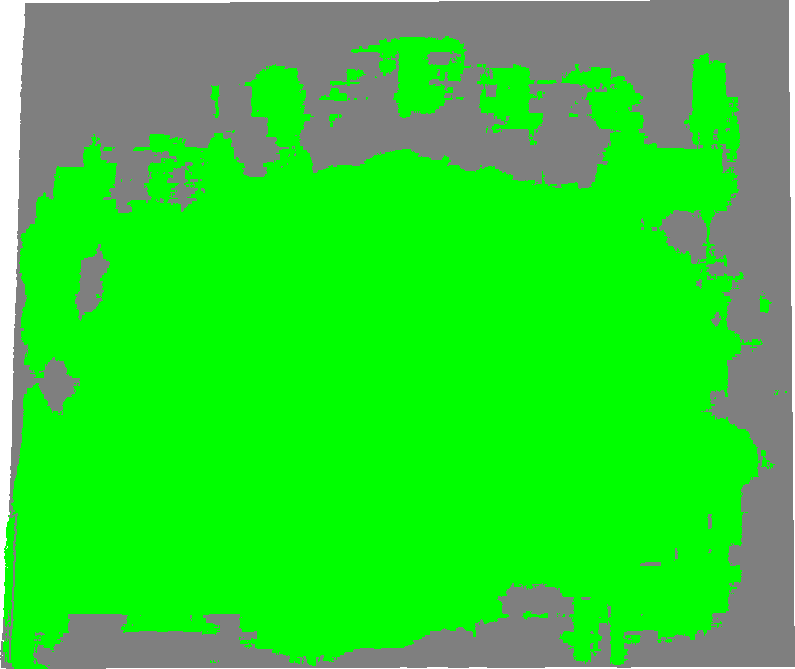}
  \end{subfigure}
  \hspace{0.05\linewidth}
  \begin{subfigure}
   \centering
    \includegraphics[width=0.45\linewidth]{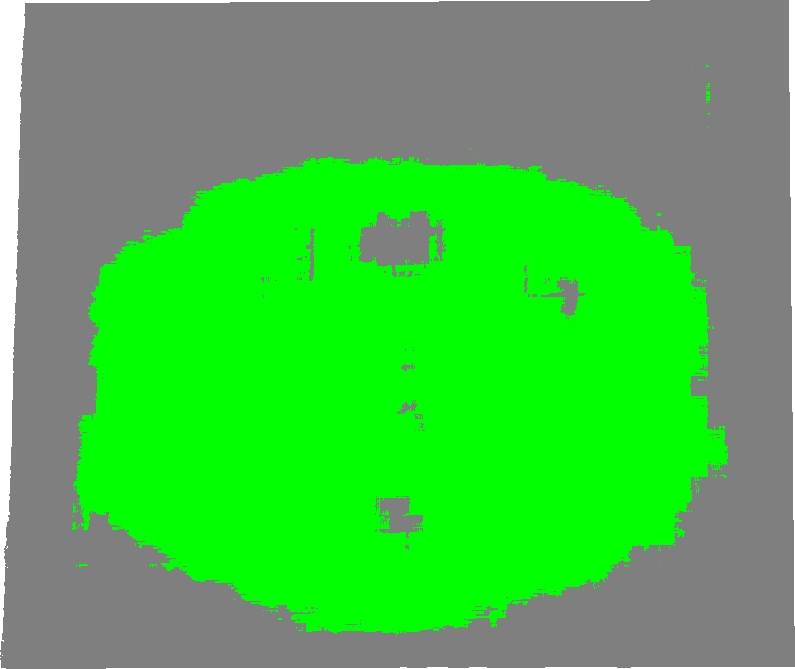}
  \end{subfigure}
  
  \caption{Prediction comparison between the network trained without (left column) and with (right column) noise signals of samples A,B,C and D (from top to bottom).}
  \label{fig:samples}
\end{figure}

However, the most interesting fact is to see if the knowledge learnt by the network is applicable to real scan data.
For this experiment, four different surface samples were modeled, each one with a dent of different size (Table~\ref{tab:samples}) following the shape of Eq.~(\ref{eq:dentFunction}), then 3D printed as $\SI[product-units=single]{15 x 15}{\centi\meter}$ tiles, sanded and painted.
\begin{table}[!h]
\centering
\begin{tabular}{|c|c|c|c|}
\hline
Sample & Length (mm) & Width (mm) & Depth (mm) \\ \hline
A & 60 & 40 & 2 \\ \hline
B & 120 & 100 & 2 \\ \hline
C & 100 & 80 & 3 \\ \hline
D & 120 & 80 & 1 \\ \hline
\end{tabular}
\caption{Dimensions of dent samples.}
\label{tab:samples}
\end{table}
The samples were scanned using the same apparatus used to acquire the noise signals above.
The comparison of the output predictions by means of the network trained without and with the noise signals is showed in Fig.~\ref{fig:samples}. Regions predicted as dented are in green. The average processing time was only $\SI{0.74}{\second}$ for $\SI{481280}{}$ points (Intel i7-7500U CPU, 12 GB of RAM).

Although no rigorous ground truth is available for these samples, the network trained using the noise signals was able to correctly segment the dented areas almost completely. The network trained using only Gaussian noise, instead, performed poorly in all the cases.

As it happens for human engineers, the identification of the dent boundaries remains a challenging task. In fact, the samples B and D produced less than clear boundaries.  This seems to confirm the thesis that, even for artificial networks, the wider and longer the dent, the more difficult its detection~\cite{CAA_CookThesis}.
The sample D, in particular, presented a maximum depth of only $\SI{1}{\milli\meter}$ and the higher width/depth ratio, hence its segmentation is harder and some false negative spots are clearly visible.
Some minor false positive spots are also noticeable in sample A and D.
Nevertheless, for the most part, the proposed method correctly distinguished dents from the undamaged surface with evident difference of performance compared to the first approach not involving the acquired noise signals.

\section{Conclusions}
The automation of aircraft dent inspection can count on modern and accurate 3D scanning technologies to acquire raw data in the form of point clouds. Without comprehensive automation, however, the task of analysing them is left to the operator.
As such, point cloud segmentation plays a fundamental role for the future of automated inspections.
Dent segmentation, in particular, is difficult because, similarly to real world inspections, the dent does not present distinctive features. Another issue is the lack of a substantial dataset containing annotated point clouds of dented surfaces.

The proposed method reduces the problem to 2D segmentation by the fitting of a quadratic function. The matrix of residuals is then fed to a fully convolutional network, with an overall processing speed of over $\SI{500000}{}$ points per second on an Intel i7-7500U CPU.
The lack of a suitable dataset was addressed by using virtual data. To prepare the network for real case scenarios, noise signals were acquired by the chosen scanning apparatus and added to the dataset during training. Thus, the network effectively learns to filter out the scanner noise and is able to correctly predict dents in real scan data, even if trained with virtual samples only.

Once the dent is correctly segmented, further post-processing algorithms may be applied to extract measures and show a full report to the engineer for the final damage evaluation.
When available, the use of a labelled dataset of real dent scans could be used for fine tuning of the model, improving performance and robustness for the effective field application.

\bibliographystyle{ieeetr} 
\bibliography{BIBLIOGRAPHYSEGMENTATION}

\section{Contact Information}
Pasquale Lafiosca \newline
pasquale.lafiosca@cranfield.ac.uk

\end{document}